\documentclass[10pt,twocolumn,letterpaper]{article}

\usepackage{iccv}
\usepackage{times}
\usepackage{epsfig}
\usepackage{graphicx}
\usepackage{amsmath}
\usepackage{amssymb}
\usepackage{blindtext}
\usepackage{multirow}
\usepackage{array}

\usepackage{xcolor}

\usepackage{caption}
\usepackage[pagebackref=true,breaklinks=true,letterpaper=true,colorlinks,bookmarks=false]{hyperref}

\iccvfinalcopy 

\def\iccvPaperID{} 

\ificcvfinal\fi
\begin{document}

\title{Dress like a Star: Retrieving Fashion Products from Videos}

\author{Noa Garcia\\
Aston University, UK\\
{\tt\small garciadn@aston.ac.uk}
\and
George Vogiatzis\\
Aston University, UK\\
{\tt\small g.vogiatzis@aston.ac.uk}
}

\maketitle

\begin{abstract}
This work proposes a system for retrieving clothing and fashion products from video content. Although films and television are the perfect showcase for fashion brands to promote their products, spectators are not always aware of where to buy the latest trends they see on screen. Here, a framework for breaking the gap between fashion products shown on videos and users is presented. By relating clothing items and video frames in an indexed database and performing frame retrieval with temporal aggregation and fast indexing techniques, we can find fashion products from videos in a simple and non-intrusive way. Experiments in a large-scale dataset conducted here show that, by using the proposed framework, memory requirements can be reduced by 42.5X with respect to linear search, whereas accuracy is maintained at around 90\%.
\end{abstract}

\section{Introduction}

Films and TV shows are a powerful marketing tool for the fashion industry, since they can reach thousands of millions of people all over the world and impact on fashion trends. Spectators may find clothing appearing in movies and television appealing and people's personal style is often influenced by the multimedia industry. Also, online video-sharing websites, such as YouTube\footnote{https://www.youtube.com/}, have millions of users generating billions of views every day\footnote{https://www.youtube.com/yt/press/statistics.html} and famous \textit{youtubers}\footnote{Users who have gained popularity from their videos on YouTube} are often promoting the latest threads in their videos.

Fashion brands are interested in selling the products that are advertised in movies, television or YouTube. However, buying clothes from videos is not straightforward. Even when a user is willing to buy a fancy dress or a trendy pair of shoes that appear in the latest blockbuster movie, there is often not enough information to complete the purchase. Finding the item and where to buy it is, most of the times, difficult and it involves time-consuming searches.

To help in the task of finding fashion products that appear in multimedia content, some websites, such as \textit{Film Grab}\footnote{http://filmgarb.com/} or \textit{Worn on TV}\footnote{https://wornontv.net}, provide catalogs of items that can be seen on films and TV shows, respectively. These websites, although helpful, still require some effort before actually buying the fashion product: users need to actively remember items from videos they have previously seen and navigate through the platform until they find them. On the contrary, we propose an effortless, non-intrusive and fast computer vision tool for searching fashion items in videos.

\begin{figure}
\centering
\includegraphics[width=0.48\textwidth]{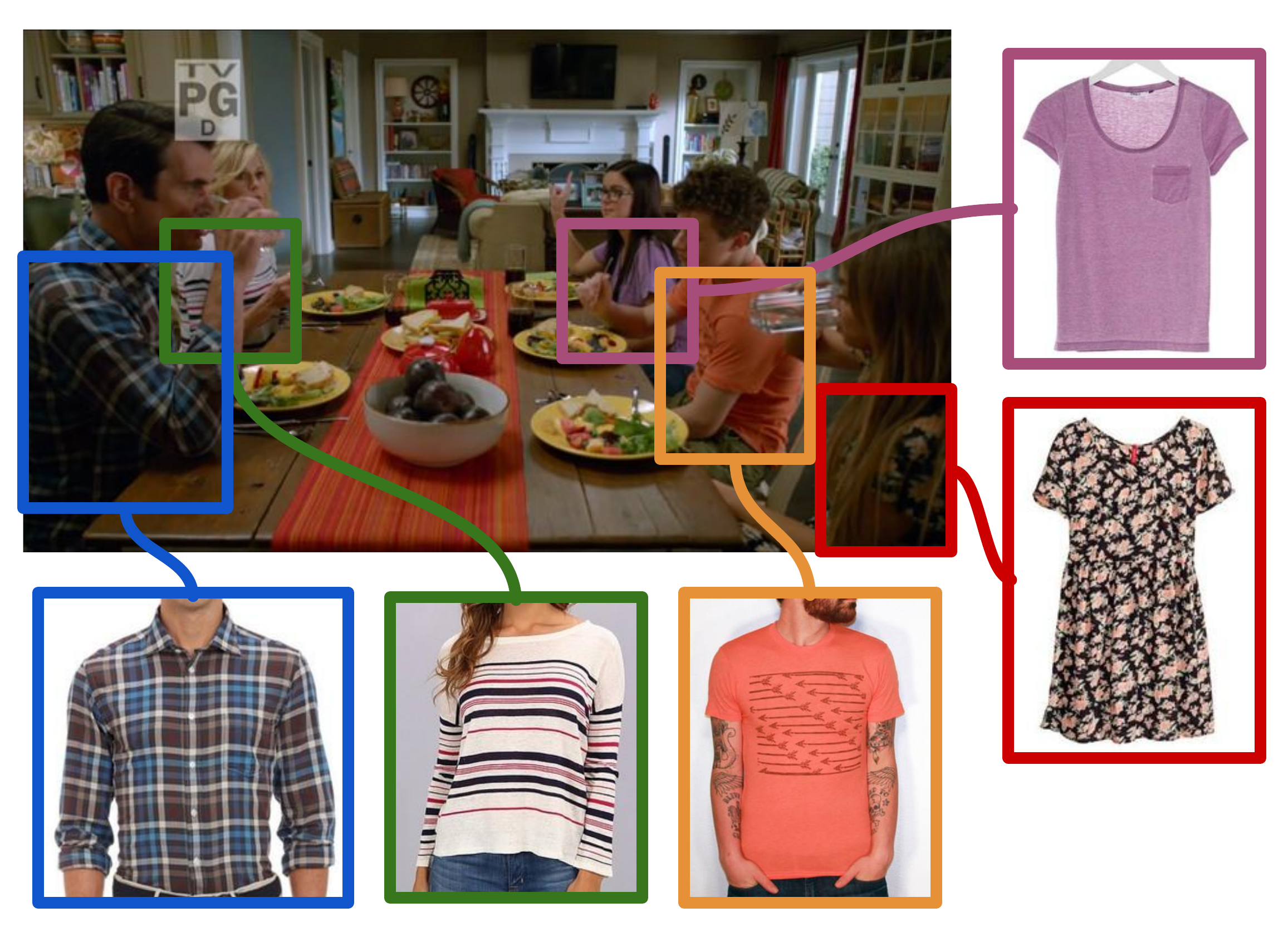}
\caption{Fashion items in a popular TV show.}
\label{fig:example}
\end{figure}

This work proposes a system for retrieving clothing products from large-scale collections of videos. By taking a picture of the playback device during video playback (i.e. an image of the cinema screen, laptop monitor, tablet, etc.), the system identifies the corresponding frame of the video sequence and returns that image augmented with the fashion items in the scene. In this way, users can find a product as soon as they see it by simple taking a photo of the video sequence where the product appears. Figure \ref{fig:example} shows a frame of a popular TV show along with its fashion items.

Recently, many computer vision applications for clothing retrieval \cite{liu2012street, hadi2015buy} and style recommendation \cite{veit2015learning, simo2015neuroaesthetics} have been proposed. Our system is close to clothing retrieval approaches in that the aim is to retrieve a product from an image. Instead of retrieving products directly as in standard clothing retrieval, we propose to first retrieve frames from the video collection. There are several reasons for that. Firstly, in standard clothing retrieval users usually provide representative images of the object of interest (e.g.  dresses in front view, high-heeled shoes in side view, etc.). In a movie, the view of the object of interest cannot be chosen, and items might be partially or almost completely occluded, such as the red-boxed dress in Figure \ref{fig:example}. Secondly, standard clothing retrieval requires to select a bounding box around the object of interest. This is undesirable in clothing video retrieval as it may distract user's attention from the original video content. Finally, performing frame retrieval instead of product retrieval in videos allows users to get the complete list of fashion items in a scene, which usually matches the character's style, including small accessories such as earrings, watches or belts. Some of these items are usually very small, sometimes almost invisible, and would be impossible to detect and recognize using standard object retrieval techniques.

Retrieving frames from videos is a challenging task in terms of scalability. An average movie of two hours duration may contain more than 200,000 frames. That means that with only five movies in the video dataset, the number of images in the collection might be over a million. To overcome scalability issues, we propose the combination of two techniques. Firstly, frame redundancy is exploited by tracking and summarizing local binary features \cite{Calonder2010,Rublee2011} into a \emph{key feature}. Secondly, key features are indexed in a kd-tree for fast search of frames. Once the scene the query image belongs to is identified, the clothing items associated to that scene are returned to the user. 

The contributions of this paper are:
\begin{itemize}
\setlength\itemsep{0em}
\item Introduction of the video clothing retrieval task and collection of a dataset of videos for evaluation.
\item Proposal of a fast and scalable video clothing retrieval framework based on frame retrieval and indexing algorithms.
\item Exhaustive evaluation of the framework, showing that similar accuracy to linear search can be achieved while the memory requirements are reduced by a factor of 42.5.
\end{itemize}

This paper is structured as follows: related work is summarized in Section \ref{sec:relatedwork}; the details of the proposed system are explained in Section \ref{sec:method}; experiment setup and results are detailed in Section \ref{sec:experiments} and \ref{sec:results}, respectively; finally, the conclusions are summarized in Section \ref{sec:conclusions}.

\section{Related Work}
\label{sec:relatedwork}
\textbf{Clothing Retrieval.} Clothing retrieval is the field concerned with finding similar fashion products given a visual query. In the last few years, it has become a popular field within the computer vision community. Proposed methods \cite{liu2012street, wei2013style, kalantidis2013getting, hadi2015buy, huang2015cross, liu2016deepfashion} are mainly focused on matching real-world images against online catalog photos. In this cross-scenario problem, some methods \cite{wei2013style, kalantidis2013getting, liu2016deepfashion} train a set of classifiers to find items with similar attributes. Other approaches \cite{liu2012street, huang2015cross, hadi2015buy} model the differences across domains by using a transfer learning matrix \cite{liu2012street} or a deep learning architecture \cite{huang2015cross, hadi2015buy}. All these methods, however, perform clothing retrieval on static images, in which a representative and recognizable part of the item of interest is contained. In the scenario of clothing retrieval from films and television, fashion items are not always shown in a representative and recognizable way across the whole duration of the video due to object occlusions and changes in camera viewpoints. Given a snapshot of a video, an alternative solution is to first, identify the frame it belongs to and then, find the products associated to it. 

\textbf{Scene Retrieval.} Scene retrieval consists on finding similar frames or scenes from a video collection according to a query image. Early work, such as Video Google \cite{Sivic2003} and others \cite{Nister2006, chum2007scalable, chen2010dynamic}, retrieve frames from video datasets by applying image retrieval methods and processing each frame independently. More recent scene retrieval approaches use temporal aggregation methods to improve scalability and reduce memory requirements. \cite{anjulan2007object, araujo2014efficient} extract hand-crafted local features from frames, track them along time and aggregate them into a single vector. Other proposals \cite{zhu2012large, araujo2015temporal, araujo2017large} produce a single scene descriptor, which improves efficiency but involves a more difficult comparison against the static query image. Recently, Araujo \textit{et al.} \cite{araujo2017large} showed that, in scene retrieval, methods based on hand-crafted features outperform convolutional neural networks. 

Similarly to \cite{anjulan2007object}, our approach is based on the aggregation of local features along temporal tracks. Instead of using SIFT features \cite{Lowe2004}, we provide evidence that binary features \cite{Calonder2010, Rublee2011} are more appropriate for the task and that they can be easily indexed in a kd-tree for fast search.
\section{Video Clothing Retrieval Framework}
\label{sec:method}

\begin{figure*}
\centering
\includegraphics[width=0.9\textwidth]{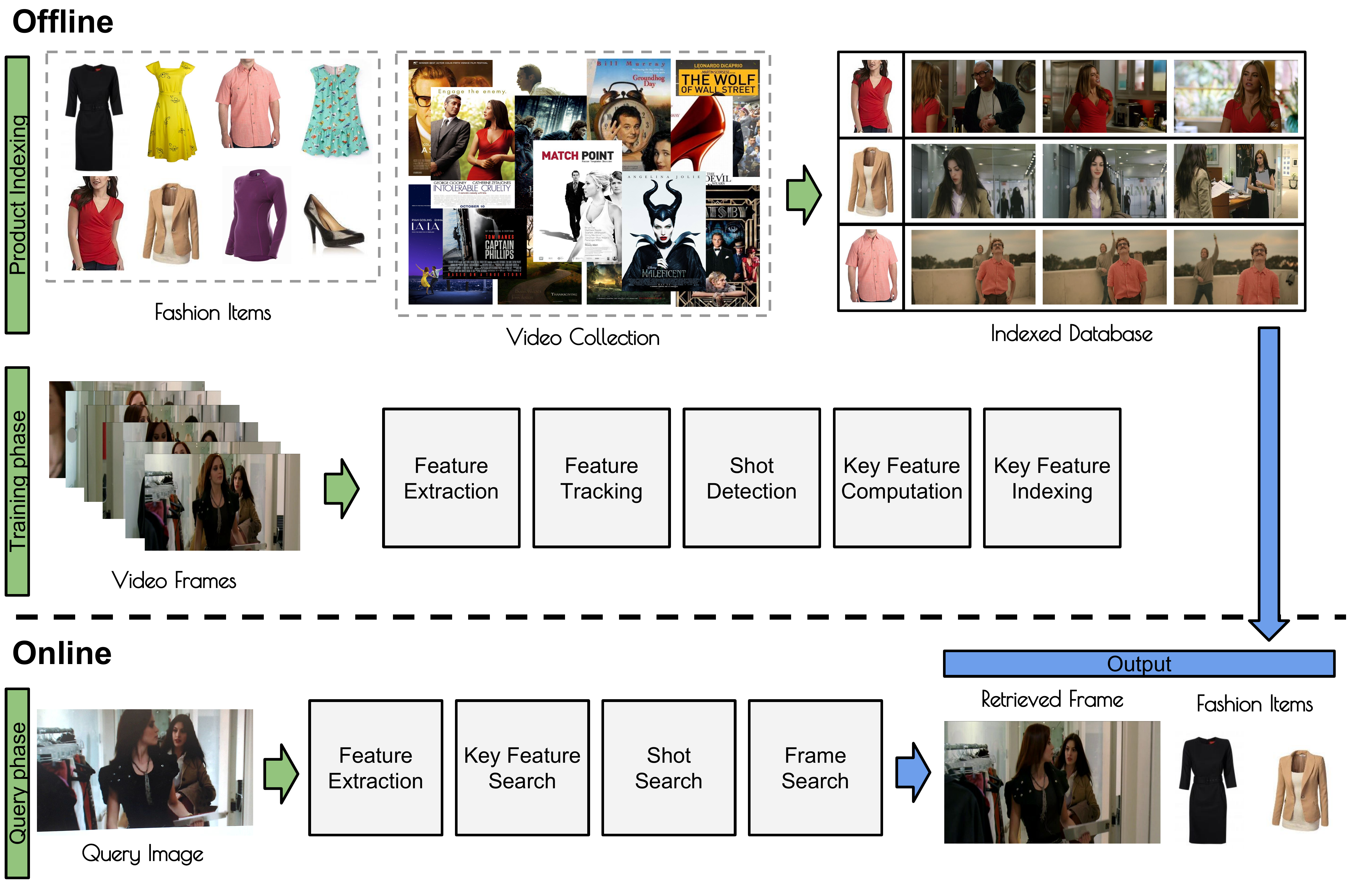}
\caption{System overview. First, during the product indexing, fashion items and frames are related in an indexed database. Then, frames are indexed by using \textit{key features}. Finally, in the query phase, query images are used to retrieve frames and find associated fashion items.}
\label{fig:BlockDiagram}
\end{figure*}

To retrieve fashion products from movies, television and YouTube, we propose a framework based on frame retrieval and fast indexing techniques. Since the number of frames explodes with the number of movies available in a collection, a key feature of this framework is its scalability. The system overview is shown in Figure \ref{fig:BlockDiagram}. There are three main modules: product indexing, training phase and query phase. The product indexing and the training phase are done offline, whereas the query phase is performed online.

In the product indexing, fashion items and frames from the video collection are related in an indexed database. This process can be done manually (e.g. with Amazon Mechanical Truck\footnote{https://www.mturk.com}) or semi-automatically with the support of a standard clothing retrieval algorithm. In the training phase, features are extracted and tracked along time. Tracks are used to detect shot boundaries and compute aggregated key features, which are indexed in a kd-tree. In the query phase, features extracted from the query image are used to find their nearest key features in the kd-tree. With those key features, the shot and the frame the query image belong to are retrieved. Finally, the fashion products associated with the retrieved frame are returned by the system.

\subsection{Feature Extraction and Tracking}
Local features are extracted from every frame in the video collection and tracked across time by applying descriptor and spatial filters. The tracking is performed in a bidirectional way so features within a track are unique (i.e. each feature can only be matched with up to two features: one in the previous frame and one in the following frame). 

\begin{figure}
\begin{tabular}{cc}
\includegraphics[height=6.8cm]{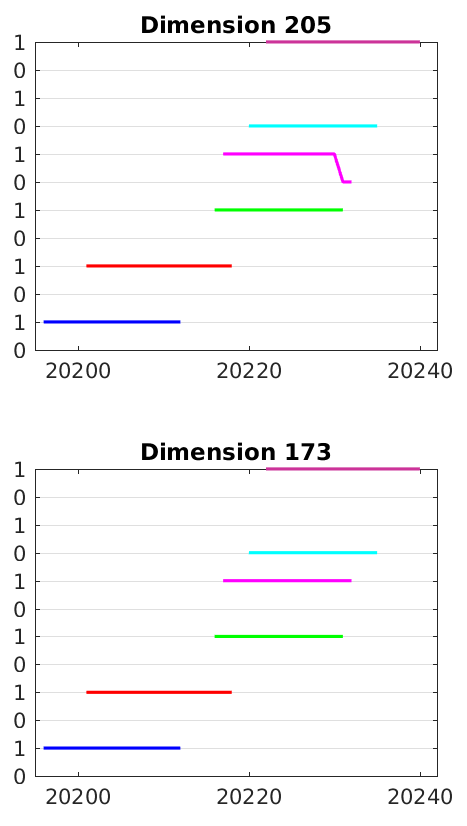} &
\includegraphics[height=6.8cm]{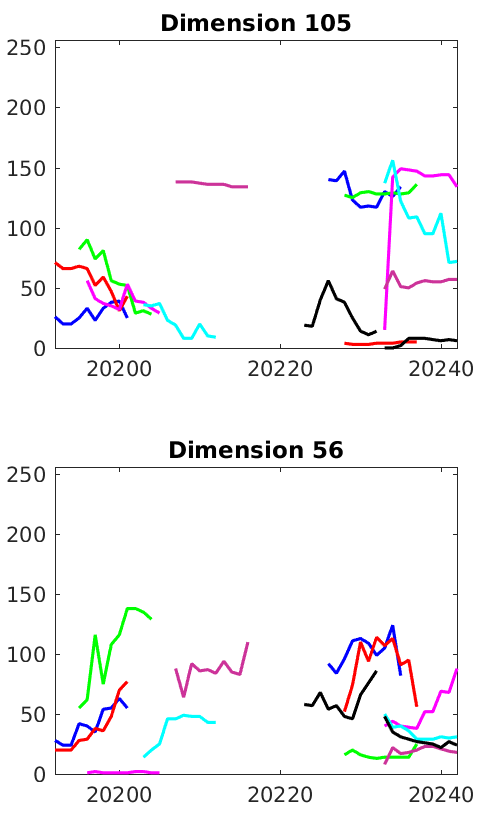}
\end{tabular}
\caption{Trajectories of sample tracks along a sequence of frames. Left: Binary features. Right: SIFT features. Binary features are more constant over time than SIFT features.}
\label{fig:tracks}
\end{figure}

As local features, instead of the popular SIFT \cite{Lowe2004}, we use binary features for two main reasons. Firstly, because Hamming distance for binary features is faster to compute than Euclidean distance for floating-points vectors. Secondly, because we find that binary features are more stable over time than SIFT, as can be seen in Figure \ref{fig:tracks}. Convolutional neural network (CNN) features from an intermediate layer of a pre-trained network were also studied. However, their results were not satisfactory as the resulting tracked features quickly started to get confused when the dataset increased.

\setlength{\tabcolsep}{3pt}
\setlength{\tabcolsep}{6pt}

\subsection{Shot Detection and Key Feature Computation}
\label{sec:shots}
Consecutive frames that share visual similarities are grouped into shots. The boundaries of different shots are detected when two consecutive frames have no common tracks. Each shot contains a set of tracks, each track representing the trajectory of a particular feature along time. We define a \textit{key feature} as the aggregation of all the features in the same track into a single vector. Subsequently, each shot is then represented by a set of key features, similarly to how frames are represented by a set of features. For each track, a key feature is computed by using majorities \cite{grana2013fast}. 

\subsection{Key Feature Indexing}
A popular method for searching in binary feature space is FLANN \cite{Muja2012}. FLANN uses multiple, randomly generated hierarchical structures and it is not obvious how it can be deployed across multiple CPUs in order to provide the scalability we are looking for. On the other hand, kd-trees are a formalism that has been shown to be highly parallelizable in \cite{Aly2011}. In this work, we modify the basic kd-tree to handle binary features and index our key features. 

In a kd-tree each decision node has an associated dimension, a splitting value and two child nodes. If the value of a query vector in the associated dimension is greater than the splitting value, the vector is assigned to the left child. Otherwise, the vector is associated to the right child. The process is repeated at each node during the query phase until a leaf node is reached. In our case, as we are dealing with binary features, each decision node has an associated dimension, $dim$, such that all query vectors, $v$, with $v[dim] = 1$ belong to the left child, and all vectors with $v[dim] = 0$  belong to the right child. The value $dim$ is chosen such that the training data is split more evenly in that node, i.e. its entropy is maximum. Note that this criterion is similar to the one used in the ID3 algorithm \cite{quinlan1986induction} for the creation of decision trees, but where the splitting attribute is chosen as the one with smallest entropy. Leaf nodes have as many as $S_L$ indices pointing to the features that ended up in that node. A first-in first-out (FIFO) queue keeps record of the already visited nodes to backtrack $B$ times and explore them later. We use a FIFO queue to ensure that even if some of the bits in a query vector are wrong, the vector can reach its closest neighbours by exploring unvisited nodes latter. With the FIFO queue we first explore the closest nodes to the root, because the corresponding bits exhibit more variability by construction of the tree.

\subsection{Search}
In the query phase, binary features are extracted from an input image and assigned to its nearest set of key features by searching down the kd-tree. Each key feature votes for the shot it belongs to. The set of frames contained in the most voted shot are compared against the input image by brute force, i.e. distances between descriptors in the query image and descriptors in the candidate frames are computed. The frame with minimum distance is retrieved. Shots are commonly groups of a few hundreds of frames, thus the computation can be performed very rapidly when applying the Hamming distance. Finally, all the fashion vectors associated with the retrieved frame are returned to the user.

\setlength{\tabcolsep}{2pt}
\begin{figure}
\centering
\begin{tabular}{ccc}
\raisebox{-.25\height}{\includegraphics[width=0.15\textwidth]{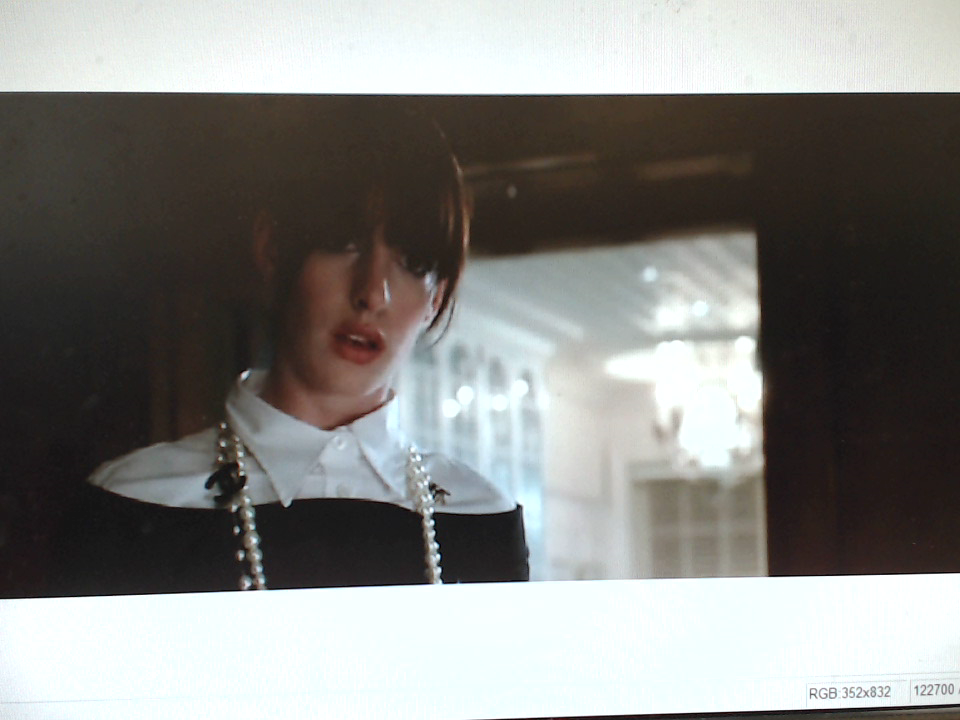}}&
\includegraphics[width=0.15\textwidth]{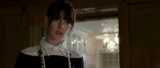}&
\includegraphics[width=0.15\textwidth]{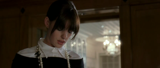}\\
\end{tabular}
\caption{Visual similarities between frames. Left: Query image. Middle: Ground truth frame. Right: Retrieved frame similar to ground truth frame, thus, Visual Match.}
\label{fig:visualMatch}
\end{figure}
\setlength{\tabcolsep}{6pt}

\section{Experiments}
\label{sec:experiments}
\textbf{Experimental Details.} To evaluate our system a collection of 40 movies with more than 80 hours of video and up to 7 million frames is used. Query images are captured by a webcam (Logitech HD Pro Webcam C920) while movies are being played on a laptop screen. To provide a ground truth for our experiments, the frame number of each captured image is saved in a text file. Movies and queries have different resolutions and aspect ratios, thus all the frames and images are scaled down to 720 pixels in width. Binary features are computed by using ORB detector \cite{Rublee2011} and BRIEF extractor \cite{Calonder2010}. In the tracking module, only matches with a Hamming distance less than 20 and a spatial distance less than 100 pixels are considered, whereas in the key feature computation algorithm, only tracks longer than 7 frames are used. The default values for the kd-tree are set at $S_L = 100$ and $B = 50$.

\textbf{Evaluation criteria.} The aim of the system is to find the fashion products in the query scene, so we only retrieve one frame per query image. The returned frame may not be the same frame as the one in the annotated ground truth but one visually similar. As long as the retrieved frame shares strong similarities with the ground truth frame, we consider it as a \emph{Visual Match}, as shown in Figure \ref{fig:visualMatch}. To measure the visual similarities between two frames, we match SURF \cite{Bay2006} features. The linear comparison between two frames that do not present any noise or perspective distortion is an easy task that almost all kinds of features can perform correctly. If the matching score between SURF features of two dataset frames is greater than a threshold, $\tau$, they are considered to be visually similar.

\setlength{\tabcolsep}{2pt}
\begin{figure}
\centering
\begin{tabular}{cc}
\raisebox{-.47\height}{\includegraphics[width=0.23\textwidth]{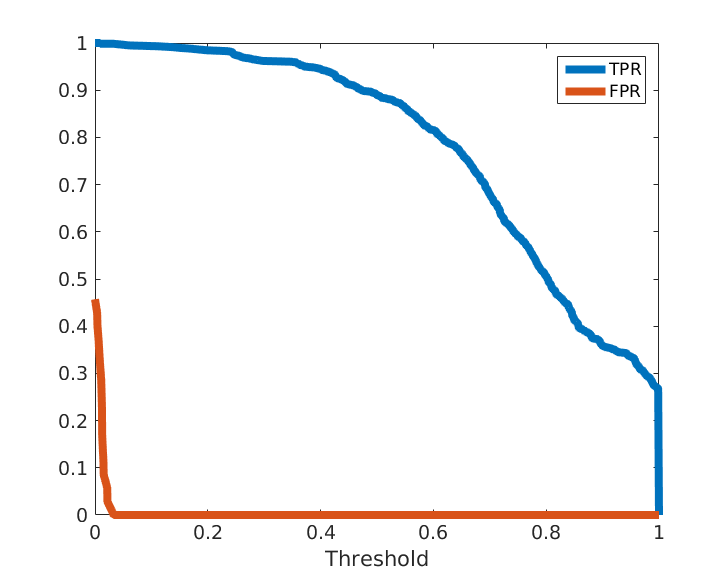}} &
\begin{tabular}{cc}
\centering
\includegraphics[width=0.11\textwidth]{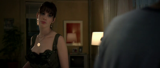}&
\includegraphics[width=0.11\textwidth]{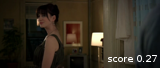}\\
\includegraphics[width=0.11\textwidth]{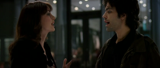}&
\includegraphics[width=0.11\textwidth]{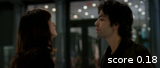}\\
\includegraphics[width=0.11\textwidth]{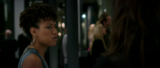}&
\includegraphics[width=0.11\textwidth]{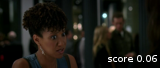}\\
\end{tabular} \\
(a) & (b)
\end{tabular}
\caption{Precision of the evaluation method: (a) TPR and FPR curves; (b) Three ground truth frames (left column) and retrieved frames (right column) along with their score.}
\label{fig:groundTruthROC}
\end{figure}
\setlength{\tabcolsep}{6pt}

To measure the precision of this evaluation method, we manually annotate either if a pair of frames (i.e. ground truth frame and retrieved frame) is a Visual Match or not, along with its score. For different values of $\tau$ the True Positive Rate (TPR) as well as the False Positive Rate (FPR) are computed as:
\begin{equation}
TPR = {TP \over TP + FN}
\qquad
FPR = {FP \over FP + TN}
\end{equation} where $TP$ is the number of true positives (Visual Match with score $> \tau$), $FP$ is the number of false positives (No Match with score $> \tau$), $TN$ is the number of true negatives (No Match with score $\leq \tau$) and $FN$ is the number of false negatives (Visual Match with score $\leq \tau$). Figure \ref{fig:groundTruthROC} shows both the TPR and the FPR computed with the annotations of 615 pairs of frames and 406 different values of $\tau$. Among all the possible values, we chose $\tau = 0.15$, with $TPR = 0.98$ and $FPR = 0$. Finally, the accuracy for a set of query images is computed as: \begin{equation}
\text{Acc} = \frac{\text{No. Visual Matches}}{\text{Total No. Queries}}
\end{equation}

\section{Experimental Results}
\label{sec:results}
To evaluate the proposed framework, two different experiments are performed. In the first one, our frame retrieval system is compared against other retrieval systems. In the second one, the performance of the frame retrieval method when scaling up the video collection is evaluated in terms of accuracy.

\subsection{Retrieval Performance}
\begin{table}
\begin{center}
\begin{tabular}{cc|>{\centering\arraybackslash}m{30pt}|>{\centering\arraybackslash}m{30pt}|>{\centering\arraybackslash}m{30pt}|>{\centering\arraybackslash}m{30pt}|>{\centering\arraybackslash}m{3pt}}
\cline{3-6}
 & & \textbf{BF} & \textbf{KT} & \textbf{KF} & \textbf{Ours} & \\[5pt]
\cline{1-6}
\cline{1-6}
\multicolumn{2}{|c|}{\begin{tabular}{@{}c@{}}\textbf{Indexed}\\\textbf{Features}\end{tabular}} & 85M & 85M & 25M & 2M & \\[15pt]
\cline{1-6}
\multicolumn{2}{|c|}{\textbf{Memory}} & 2.53GB & 2.53GB & 762MB & 61MB & \\[15pt]
\cline{1-6}
\multicolumn{1}{ |c| }{\parbox[t]{2mm}{\multirow{4}{*}{\rotatebox[origin=r]{90}{\textbf{Accuracy}}}}} & B = 10 & \multirow{4}{*}{0.98}	& 0.90 & 0.91 & 0.92 & \\[5pt]
\cline{2-2}\cline{4-6}
\multicolumn{1}{ |c| }{} & B = 50 & 	& 0.94 & 0.92 & 0.93 & \\[5pt]
\cline{2-2}\cline{4-6}
\multicolumn{1}{ |c| }{} & B = 100 & 	& 0.96 & 0.93 & 0.94 & \\[5pt]
\cline{2-2}\cline{4-6}
\multicolumn{1}{ |c| }{} & B = 250 & 	& 0.97 & 0.93 & 0.94 & \\[5pt]
\cline{1-6}
\end{tabular}
\end{center}
\caption{Comparison between different systems on a single movie. Accuracy is shown for four backtracking steps, B. Our system achieves comparable results by using 42.5x less memory than BF and KT and 12.5x less memory than KF.}
\label{table:comparision}
\end{table}

\begin{table*}
\begin{center}
\begin{tabular}{ l | c  | c  | c  | c  | c  | c }
\textbf{Title} & \textbf{N. Frames} & \textbf{N. Features} & \textbf{N. Shots} & \textbf{N. Key Features} & \textbf{N. Queries} & \textbf{Accuracy} \\
\hline
The Help & 210387 & 101M & 1726 & 2.2M & 813 & 0.98\\
Intolerable Cruelty & 179234 & 86M & 1306 & 2M & 544	& 0.97 \\
Casablanca & 147483 & 71M & 881 & 1.5M & 565	&  0.96\\
\hline
Witching \& Bitching & 163069 & 66M & 4193 & 0.8M & 588 & 0.74 \\
Pirates of the Caribbean 3 & 241127 & 108M & 3695 & 1.7M & 881	& 0.74 \\
Captain Phillips & 190496 & 59M & 7578 & 0.6M & 618	& 0.67 \\
\hline
\textbf{Total} & \textbf{7M} & \textbf{3040M} & \textbf{116307} & \textbf{58M} & \textbf{25142} & \textbf{0.87}
\end{tabular}
\end{center}
\caption{Summary of the results from the experiments on the dataset containing 40 movies. Due to space restrictions, only three of the best and worst accuracies are shown, along with the total results for the whole database.}
\label{table:moviesSummary}
\end{table*}

First, we compare our system against other retrieval frameworks. 

\textbf{BF.} Brute Force matcher, in which query images are matched against all frames and all features in the database. Brute Force system is only used as an accuracy benchmark, since each query take, in average, 46 minutes to be processed. Temporal information is not used.

\textbf{KT.} Kd-Tree search, in which all features from all frames are indexed using a kd-tree structure, similarly to \cite{Aly2011}. Temporal information is not used.

\textbf{KF.} Key Frame extraction method \cite{sun2008video}, in which temporal information is used to reduce the amount of frames of each shot into a smaller set of key frames. Key frames are chosen as the ones at the peaks of the distance curve between frames and a reference image computed for each shot. For each key frame, features are extracted and indexed in a kd-tree structure. 

We compare these three systems along with the method proposed here, using a single movie, \textit{The Devil Wears Prada}, consisting of 196,572 frames with a total duration of 1 hour 49 minutes and 20 seconds. To be fair in our comparison, we use binary features in each method. The results of this experiment are detailed in Table \ref{table:comparision}. The performance is similar for the four systems. However, the amount of processed data and the memory requirements are drastically reduced when the temporal information of the video collection is used. In the case of our system, by exploiting temporal redundancy between frames, the memory is reduced by 42.5 times with respect to BF and KT and by 12.5 times with respect to the key frame extraction technique. Theoretically, that means that when implemented in a distributed kd-tree system as the one in \cite{Aly2011}, where the authors were able to process up to 100 million images, our system might be able to deal with 4,250 million frames, i.e. more than 20,000 movies and 40,000 hours of video. 

\begin{figure}
\centering
\includegraphics[width=0.45\textwidth]{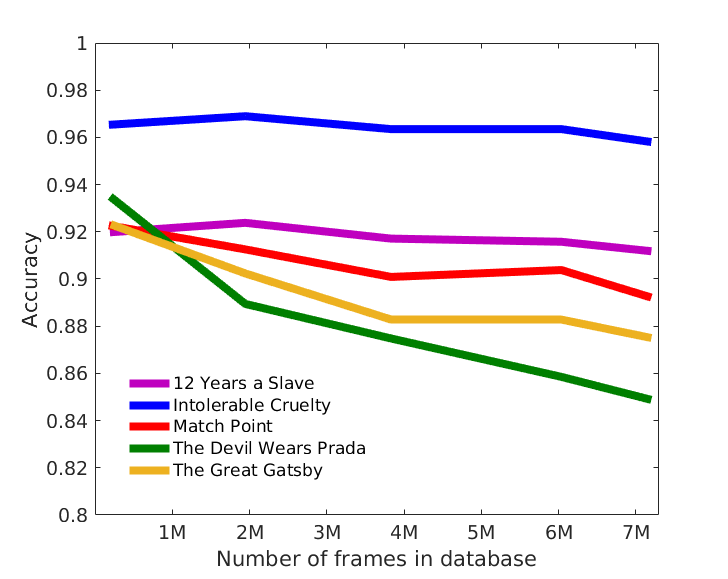}
\caption{Accuracy vs Database size for 5 different movies.}
\label{fig:AccVSDB}
\end{figure}

\subsection{Large-scale Dataset}
In this experiment, we explore the scalability of our framework by increasing the dataset up to 40 movies and 7 million frames. The collection is diverse and it contains different movie genres such as animation, fantasy, adventure, comedy or drama, to ensure that a wide range of movies can be indexed and retrieved with the proposed algorithm. 

Table \ref{table:moviesSummary} shows the results of this experiment. It can be seen that, by using our temporal aggregation method, the amount of data is reduced from 7 million frames and 3,040 million features to only 116,307 shots and 58 million key features. Even so, the total number of key features in the 40 movie collection is still smaller than the 80 million features that, in average, a single movie contains. The total accuracy over the 40 movies is 0.87, reaching values of 0.98 and 0.97 in \textit{The Help} and \textit{Intolerable Cruelty} movies, respectively. Movies with very dark scenes such as \textit{Captain Phillips} and \textit{Pirates of the Caribbean 3} perform worst, as fewer descriptors can be found in those kinds of dimly lit images.

Figure \ref{fig:AccVSDB} shows the evolution of accuracy when the size of the database increases for five different movies. It can be seen that most of the movies are not drastically affected when the number of frames in the database is increased from 200,000 to 7 million. For example, both \textit{Intolerable Cruelty} and \textit{12 Years a Slave} movies maintain almost a constant accuracy for different sizes of the collection. Even in the worst case scenario, \textit{The Devil Wears Prada} movie, the loss in accuracy is less than a 8.5\%. This suggests that our frame retrieval system is enough robust to handle large-scale video collections without an appreciable loss in performance.

\section{Conclusions}
\label{sec:conclusions}
This work proposes a framework to perform video clothing retrieval. That is, given a snapshot of a video, identify the video frame and retrieve the fashion products that appear in that scene. This task would help users to easily find and purchase items shown in movies, television or YouTube videos. We propose a system based on frame retrieval and fast indexing. Experiments show that, by using our temporal aggregation method, the amount of data to be processed is reduced by a factor of 42.5 with respect to linear search and accuracy is maintained at similar levels. The proposed system scales well when the number of frames is increased from 200,000 to 7 million. Similar to \cite{Aly2011} our system can easily be parallelised across multiple CPUs. Therefore, the encouraging experimental results shown here indicate that our method has the potential to index fashion products from thousands of movies with high accuracy. In future work we intend to further increase the size of our dataset in order to identify the true limits of our approach. Furthermore it is our intention to make all data publicly available so that other researchers can investigate this interesting retrieval problem.

{\small
\bibliographystyle{ieee}
\bibliography{egbib}
}

\end{document}